\documentclass[runningheads]{llncs}
\usepackage{graphicx}

\usepackage{tikz}
\usepackage{comment}
\usepackage{amsmath,amssymb} 
\usepackage{color}

\usepackage[algo2e]{algorithm2e}
\usepackage{algorithm}

\begin{document}
\pagestyle{headings}
\mainmatter
\def\ECCVSubNumber{6621}  

\title{Quantum-soft QUBO Suppression for Accurate Object Detection} 

\titlerunning{Quantum-soft QUBO Suppression for Accurate Object Detection}
%
\author{Junde Li \and Swaroop Ghosh}
\authorrunning{J. Li and S. Ghosh}
%
\institute{The Pennsylvania State University, University Park, PA 16802, USA
\email{\{jul1512,szg212\}@psu.edu}}
\maketitle

\begin{abstract}
Non-maximum suppression (NMS) has been adopted by default for removing redundant object detections for decades. It eliminates false positives by only keeping the image $\mathcal{M}$ with highest detection score and images whose overlap ratio with $\mathcal{M}$ is less than a predefined threshold.  However,  this  greedy  algorithm  may  not  work  well  for  object  detection under occlusion scenario where true positives with lower detection scores are possibly suppressed.  In this paper, we first map the task of removing redundant detections into Quadratic Unconstrained Binary Optimization (QUBO) framework that consists of detection score from each bounding box and overlap ratio between pair of bounding boxes. Next, we solve the QUBO problem using the proposed Quantum-soft QUBO Suppression (QSQS) algorithm for fast and accurate detection by exploiting quantum computing advantages. Experiments  indicate  that  QSQS  improves  mean average precision from 74.20\% to 75.11\% for PASCAL VOC 2007. It consistently outperforms NMS and soft-NMS for \textit{Reasonable} subset of benchmark pedestrian detection CityPersons.
\end{abstract}

\section{Introduction}
Object detection helps image semantic understanding by locating and classifying objects in many applications such as, image classification \cite{imagenet}, face recognition \cite{face2}, \cite{face}, autonomous driving \cite{deepdriv}, \cite{autodri}, and surveillance \cite{surveillance}. It has been developed from handcrafted features to Convolutional Neural Network (CNN) features, and from sliding windows to region proposals. Redundant bounding boxes are eliminated by default using Non-Maximum Suppression (NMS) for decades. However, greedy NMS can lead to detection misses for detection of partially occluded objects by completely suppressing neighbouring bounding boxes when overlap threshold is reached. While improvements have been made in \cite{soft-nms}, \cite{yihui}, \cite{dpp}, \cite{collins}, a better algorithm is still a necessity for accurately retaining true positive object locations, especially for pedestrian detection where occluded people are common. This is particularly important for autonomous systems that need to make accurate decisions even under occluded situations to guarantee safe operations.

Quadratic Unconstrained Binary Optimization (QUBO) framework has been used for removing redundant object detections in \cite{collins}, but neither its detection accuracy in mAP metric nor detection delay are as good as traditional non-maximum suppression \cite{soft-nms}, \cite{yihui}. While no further research on QUBO for object detection has been performed since \cite{collins}, its potential for accurate object detection by incorporating multiple detection scores and overlap ratios is worth being explored. Importantly, QUBO problem can be efficiently solved by quantum computing than any classical algorithms. Recent advances in quantum computing can be exploited to address this issue.

\begin{figure}
\centering
\includegraphics[width=8cm]{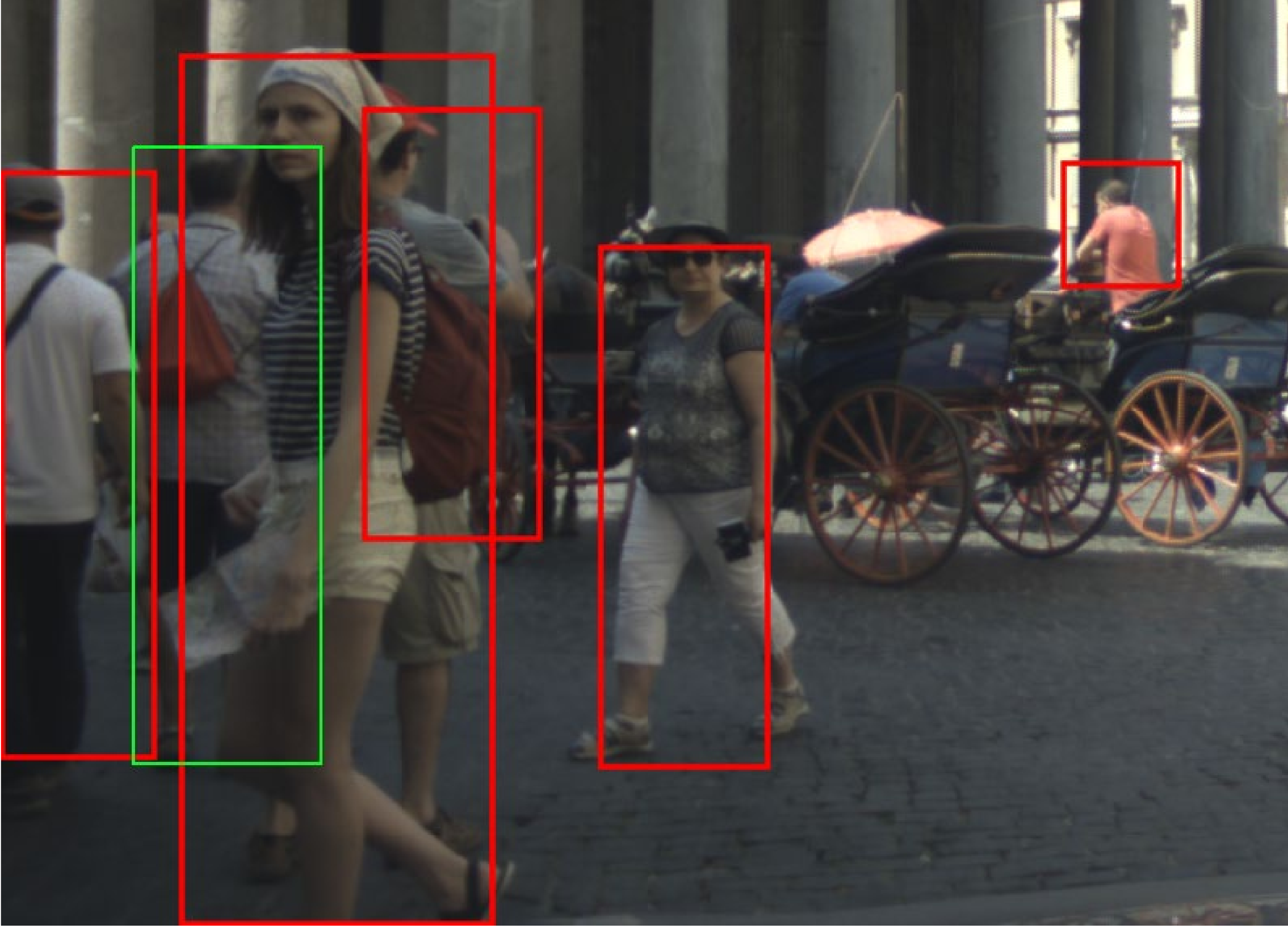}
\caption{Pedestrian detection using Non-Maximum Suppression (NMS) for removing false positive detections. Detection indicated by green box is eliminated under harsh suppression threshold due to significant overlap with front detection.}
\label{occlud}
\end{figure}

Quantum advantage (supremacy) has been specifically demonstrated in \cite{sup}, \cite{sup-comput}, and practically in applications of flight gate scheduling \cite{flight}, machine learning \cite{sup-machine}, and stereo matching \cite{stereo}. In this study, we focus on another real-world application of removing redundant object detections using quantum annealer (QA). Quantum annealing system first embeds binary variables in QUBO framework to physical qubits, and outputs optimized binary string indicating whether corresponding bounding boxes to be retained or removed. For example, D-Wave 2000Q quantum annealer supports embedding of up to 2048 qubits which is sufficient for object detection task of this work.

\begin{figure*}
\centering
\includegraphics[width=12cm]{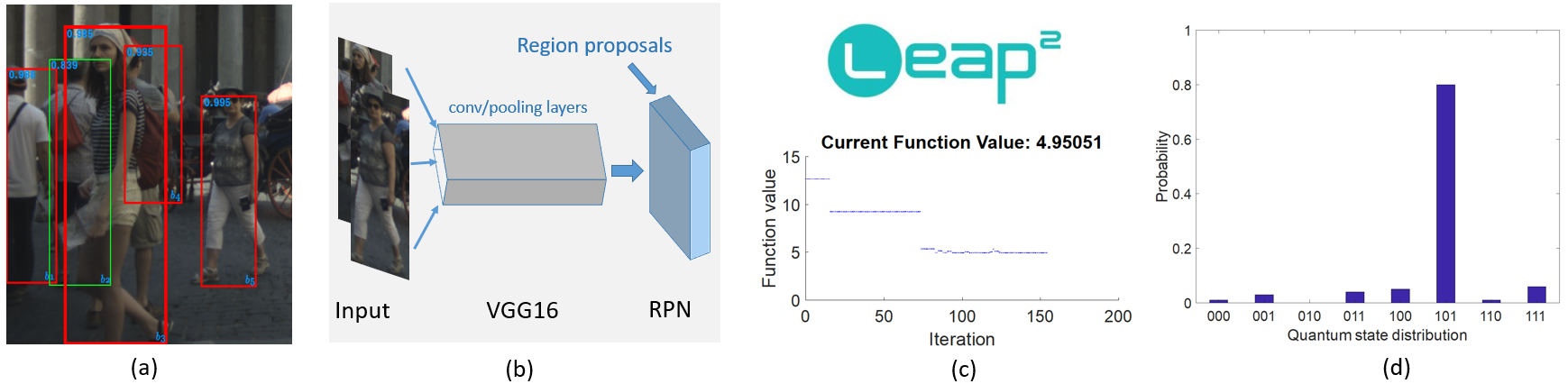}
\caption{Formulation and solution steps of QUBO problem for removing redundant detections using our proposed QSQS algorithm: (a) generate detection score from detector to serve as linear term of QUBO model (boxes are initially filtered by non-maximum suppression); (b) obtain regions of interest through a series of conv and pooling layers and Region Proposal Network (RPN); (c) use quantum annealing process through D-Wave LEAP service for filtering out false detections; (d) retrieve (near-)optimal solution of QUBO problem through quantum state readout (for example bit string '101' indicates only first and third bounding boxes are kept).}
\label{flow}
\end{figure*}

NMS recognizes objects less well for crowded and partially occluded scenes as occluded objects are likely to be completely suppressed due to significant overlap with front objects that have higher detection scores (Fig. \ref{occlud}). QUBO framework was developed for addressing this challenge \cite{collins}, with its linear term modeling detection score of each bounding box $b_i$ and quadratic term (aka pairwise term) modeling overlap ratio of each pair of bounding boxes $or(b_i, b_j)$. However, QUBO framework indicated degraded accuracy than the standard NMS \cite{yihui} from a generic detection perspective. In order to improve its performance, we propose Quantmum-soft QUBO Suppression (QSQS) by adapting the idea of Soft-NMS \cite{soft-nms} for decaying classification score of $b_i$ which has high overlap with target detection $\mathcal{M}$ (i.e., detection with highest score). Besides, we further improve its performance by incorporating spatial correlation features \cite{dpp} as additional metric of overlap ratio, apart from metric of Intersection Over Union (IOU). Finally, the QUBO problem is solved using real D-Wave 2000Q quantum system with unmatched computing efficiency compared to classical solvers. The overall detection framework using soft-QUBO and quantum annealing is illustrated in Fig. \ref{flow}. The contributions of this paper are three-fold: (1) We propose an novel hybrid quantum-classical algorithm, QSQS, for removing redundant object detections for occluded situations; (2) We implement the hybrid algorithm on both GPU of classical computer for running convolutional neural network, and QPU (quantum processing unit) of quantum annealer for harnessing quantum computing advantage; (3) The proposed QSQS improves mean average precision (mAP) from 74.20 to 75.11 percent for generic object detection, and outperforms NMS and soft-NMS consistently for pedestrian detection on CityPersons \textit{Reasonable} subset. However, our method takes 1.44X longer average inference time than NMS on PASCAL VOC 2007.

\section{Related Work}
\textbf{Non-Maximum Suppression:} NMS is an integral part of modern object detectors (one-stage or two-stage) for removing false positive raw detections. NMS greedily selects bounding box $\mathcal{M}$ with highest detection score, and all close-by boxes whose overlap with $\mathcal{M}$ reach hand-crafted threshold are suppressed. This simple and fast algorithm is widely used in many computer vision applications, however, it hurts either precision or recall depending on predefined parameter, especially for crowded scenes where close-by high scoring true positives are very likely to be removed. Soft-NMS \cite{soft-nms} improves the detection accuracy by decaying detection scores of close-by boxes by a linear or Gaussian rescoring function. Learning NMS \cite{learning-nms} eliminates non-maximum suppression post-processing stage so as to achieve true end-to-end training. However, the accuracy of these false positive suppression schemes are not high enough.\\

\textbf{Quadratic Unconstrained Binary Optimization (QUBO):} Quadratic binary optimization is the problem of finding a binary string vector that maximizes the objective function $C(\boldsymbol{x})$ which is composed of linear and paired terms. QUBO framework \cite{collins} has been proposed in past for suppressing false positive detections, and achieved better accuracy than NMS for pedestrian detection. However, it has been concluded that QUBO may not be optimal for other detection situations. QUBO framework is inherently suitable for crowded and partially occluded object detection as classification scores of all bounding boxes and overlaps among all pairs of them are equally considered. In this study, we aim to improve QUBO performance on detection accuracy by incorporating spatial features into pairwise terms of the cost function, and generalize its detection accuracy robustness for both crowded and non-crowded dataset.\\

\begin{figure*}\small
\centering
\includegraphics[width=12cm]{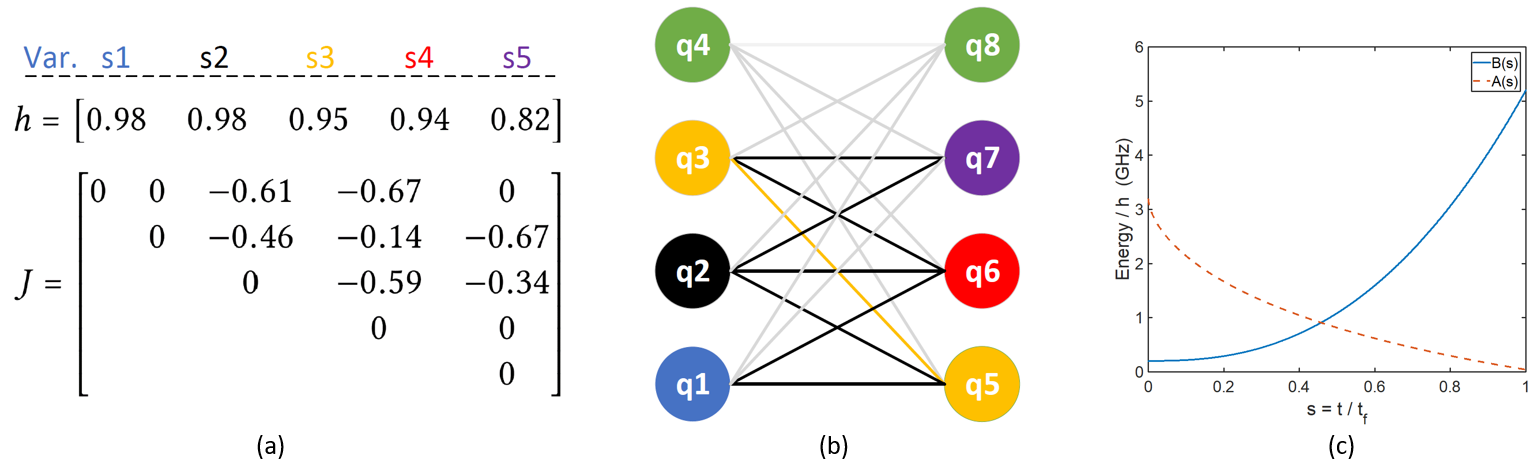}
\caption{(a) Flux biases $h$ on each qubit $s_i$ and couplings $J_{ij}$ between qubits of Ising Hamiltonian of an example object detection instance with 5 bounding boxes; (b) one optimal minor embedding of these 5 binary variables of Ising model to 6 physical qubits in one unit cell of D-Wave 2X quantum annealer (the cluster of 2 physical qubits $q_3$ and $q_5$ represents binary variable $s_3$, and variables $s_1$, $s_2$, $s_4$, and $s_5$ are represented by physical qubits $q_1$, $q_2$, $q_6$, and $q_7$, respectively); (c) typical profile of annealing functions of tunneling energy $A(s)$ and Ising Hanmiltonian energy $B(s)$ ($t_f$ is the total annealing time).}
\label{qa}
\end{figure*}

\textbf{Quantum Computing:} Quantum supremacy has been recently demonstrated by Google \cite{sup} using a quantum processor with 53 superconducting qubits. It has been shown that quantum computer is more than billion times faster in sampling one instance of a quantum circuit than a classical supercomputer. Quantum computing is first proposed for object detection using Quantum Approximation Optimization Algorithm (QAOA) implemented on a universal gate based quantum simulator \cite{jli}, \cite{ash2019qure}. However, it is only an exploratory study rather than a real-world object detection application due to gate error and limited number of qubits on gate based quantum computer. Compared to QAOA, we propose a novel hybrid quantum-classical algorithm, QSQS, specifically for object detection task using quantum annealing. In this study, the NP-hard QUBO problem is solved by quantum annealing with pronounced computing advantage, in terms of speed and accuracy, over classical algorithms such as, Tabu and Greedy Search.

\section{Approach}
This section briefly explains the redundant detection suppression task of object detection addressed by QUBO framework and its suitability as standard NMS alternative. The rationale behind the quantum advantage of quantum annealing over classical algorithms is also explained.

\subsection{Quadratic Unconstrained Binary Optimization}
QUBO is an unifying framework for solving combinatorial optimization problem which aims to find an optimal subset of objects from a given finite set. Solving such problem through exhaustive search is NP-hard as its runtime grows exponentially $\mathcal{O}(2^N)$ with number of objects $N$ in the set. Finding optimal detections to be kept from raw detections generated from object detector falls under the same category of combinatorial optimization problem. The optimal solution is a binary string vector $\boldsymbol{x} = (x_1, x_2, ..., x_N)^T$ that maximizes the objective function $C(\boldsymbol{x})$. The standard objective function is typically formulated by linear and quadratic terms as follows:

\begin{equation}\label{e1}
   C(\boldsymbol{x}) = \sum\limits_{i=1}^{N}c_ix_i + \sum\limits_{i=1}^{N}\sum\limits_{j=i}^{N}c_{ij}x_ix_j = \boldsymbol{x}^TQ\boldsymbol{x}
\end{equation}

where $\forall i, x_i \in \{0, 1\}$ are binary variables, $c_i$ are linear coefficients and $c_{ij}$ are quadratic coefficients. Each object detection instance is represented by a specific upper triangular matrix $Q$, where $c_i$ terms make up the diagonal elements and $c_{ij}$ terms constitute off-diagonal elements of the matrix. For object detection problem of this work, $N$ denotes the number of detection bounding boxes, $c_i$ denotes the detection score of each $b_i$, and $c_{ij}$ denotes negative value of $or(b_i, b_j)$ for penalizing high overlap between pair of bounding boxes.

Theoretically, QUBO is a suitable alternative to greedy NMS for redundancy suppression as overlap $or(b_i, b_j)$ between each pair of detections is fully considered in QUBO while NMS only considers overlap $or(\mathcal{M}, b_i)$ between highest-scoring detection and others. Besides, QUBO is also able to uniformly combine multiple linear and quadratic metrics from different strategies and even detectors. Put differently, matrix  $Q$ can be formed by the sum of multiple weighted linear and pair matrices for more accurate detection. We adopt a single linear matrix $\mathcal{L}$ and two pair matrices $\mathcal{P}_1$ and $\mathcal{P}_2$ as follows:

\begin{equation}\label{e2}
   Q = w_1\mathcal{L} - w_2\mathcal{P}_1 - w_3\mathcal{P}_2
\end{equation}

where $\mathcal{L}$ is diagonal matrix of objectness scores, $\mathcal{P}_1$ is one pair matrix for measuring IOU overlap, and $\mathcal{P}_2$ is another pair matrix for measuring spatial overlap \cite{dpp}. The additional spatial overlap feature expects to enhance the detection accuracy by comparing positional similarity between each pair of the proposed regions.

\subsection{Quantum Annealing}

Quantum Annealing has been shown to be advantageous over classical algorithms in terms of computational time for QUBO problems \cite{energy}. Classical algorithm such as Tabu Search \cite{multistart} reduces time complexity from exponential to polynomial at the expense of reduced accuracy. Few problem instances with varying number of variables are run to compare the accuracy between Tabu Search and quantum annealing. Tabu Search shows 87.87\% accuracy for 9-variable problem relative to ground truth solution using brute-force, and its accuracy decreases to 66.63\% for 15-variable problem. However, quantum annealing with 1000 shots achieves the optimal strings with ground truth solutions for these problem instances. Therefore, only quantum algorithm is considered as QUBO solver in this study. 

Quantum Annealing is a metaheuristic optimization algorithm running on quantum computational devices for specially solving QUBO problems. Every QUBO instance is first mapped to its Ising Hamiltonian through equation $s_i = 2x_i - 1$ for making use of flow of the currents in the superconductor loops \cite{nasa}. The Ising Hamiltonian of an example object detection case with 5 bounding boxes is shown in Fig. \ref{qa}(a). Minor embedding is performed to map binary variables to clusters of physical qubits due to limited connectivity of D-Wave QPU architecture. Fig. \ref{qa}(b) displays one optimal minor embedding of the 5 binary variables of the Ising Hamiltonian in Fig. \ref{qa}(a). The cluster of 2 physical qubits $q_3$ and $q_5$ represents a single binary variable $s_3$, while $q_1$, $q_2$, $q_6$, and $q_7$ represent binary variables $s_1$, $s_2$, $s_4$, and $s_5$, respectively. This embedding only takes one unit cell of Chimera graph of D-Wave QPU. In Fig. \ref{qa}(b), physical qubits $q_4$ and $q_8$ are unused, and couplings between logical qubits are denoted with black edges and yellow edge denotes coupling within a chain of qubits.


Quantum annealing is evolved under the control of two annealing functions $A(s)$ and $B(s)$. The time-dependent Hamiltonian of annealing follows the equation below:
\begin{equation}\label{e4}
   H(t) = A(s)H_0 + B(s)H_1
\end{equation}
where $H_1$ is the Ising Hamiltonian of QUBO problem and $H_0$ is the initial Hamiltonian fixed in D-Wave system. As shown in Fig. \ref{qa}(c), Ising Hamiltonian energy increasingly dominates system energy in accordance with profile configuration of $A(s)$ and $B(s)$. Note, $t_f$ is the total annealing time.

\subsection{Quantum-soft QUBO Suppression}
We develop a novel hybrid quantum-classical algorithm i.e., Quantum-soft QUBO Supression (QSQS) for accurately suppressing redundant raw detection boxes. QSQS provides a means to uniformly combine multiple linear and pair terms as a QUBO problem, and harnesses quantum computational advantage for solving such problem. As shown in Algorithm \ref{qsqs}, it contains three tunable weights $w_1$, $w_2$ and $w_3$ for detection confidence, intersection over union, and spatial overlap feature (refer to \cite{dpp}), respectively.

\begin{algorithm}[H]
\SetAlgoLined
\caption{Quantum-soft QUBO Suppression (QSQS)}\label{qsqs}
\textbf{Input:} $\mathcal{B}$ - N x 4 matrix of detection boxes;
$\mathcal{S}$ - N x 1 vector of corresponding detection scores; $O_t$ - detection confidence threshold.\\
\textbf{Output:} $\mathcal{D}$ - final set of detections.\\

$\mathcal{B} = \{b_1, b_2, ..., b_N\}$, $\mathcal{S} = \{s_1, s_2, ..., s_N\}$ \\
$\mathcal{D} = \{\}$, $Q = \{\}$ \\
\For{$i \gets 1$ \KwTo $N$}{
    {$Q[i,i] \gets w_1S_i$}\\
    \For{$j \gets i$ \KwTo $N$}{
        {$Q[i,j] \gets -w_2IoU(b_i, b_j)$} \\
        {$SpatFeat \gets getSpatialOverlapFeature(b_i, b_j)$} \\
        {$Q[i,j] \gets Q[i,j]-w_3$SpatFeat} \\
    }
}

$(\mathcal{B}_{kept}, \mathcal{B}_{soft})$ $\gets quantumAnnealing(-Q)$ \\
$\mathcal{D} \gets \mathcal{D} \cup \mathcal{B}_{kept}$   \\
\For{$b_i$ in $\mathcal{B}_{soft}$}{
    {$b_m \gets argmax(IoU(\mathcal{B}_{kept}, b_i))$}\\
    {$s_i \gets s_if(IoU(b_m, b_i))$} \\
    \If{$s_i \geq O_t$}{
        {$\mathcal{D} \gets \mathcal{D} \cup b_i$} \\
    }
}
\Return{$\mathcal{D}, \mathcal{S}$}
\end{algorithm}

QUBO problems are formulated from initial detection boxes $\mathcal{B}$ and corresponding scores $\mathcal{S}$. We first negate all elements in square matrix $Q$ of the QUBO instance to convert original maximization to a minimization problem to match with the default setting of the D-Wave system. Then a binary quadratic model is formed using negated diagonal and off-diagonal entries in matrix $Q$ through D-Wave cloud API \texttt{dimod}. Optimal or near-optimal binary string vector $\boldsymbol{x} = (x_1, x_2, ..., x_N)^T$ is the string vector with highest sampling frequency after 1000 quantum state readouts. Value of 1 or 0 for each $x_i$ indicates keep or removal of corresponding detection $b_i$.

Instead of removing all boxes $\mathcal{B}_{soft}$ returned from the annealer, a rescoring function $f(IoU(b_m, b_i))$ is applied for mitigating probability of detection misses. The Gaussian weighting function \cite{soft-nms} is as follows:
\begin{equation}\label{e5}
   s_i = s_iexp(-\frac{IoU(b_m, b_i)^2}{\sigma}), \forall b_i \in \mathcal{B}_{soft}
\end{equation}

where $\sigma$ is fixed at 0.5 in this study. Average precision drop caused by picking only highest scoring detection may be avoided through this soft rescoring step.

\section{Experiments}
In this section, we first introduce evaluation datasets and metrics. Next, we describe the implementation details followed by ablation study of our approach. Finally, the comparison with the state-of-the-art is presented. All experiments and analyses are conducted on two application scenarios, i.e., generic object detection and pedestrian detection, to validate the proposed quantum-soft QUBO suppression method.

\subsection{Datasets and Evaluation Metrics}
\subsubsection{Datasets.} We choose two datasets for generic object detection evaluation. The PASCAL VOC 2007 \cite{voc-2007} dataset has 20 object categories and 4952 images from test set, while the large-scale MS-COCO \cite{coco} dataset contains 80 object categories and 5000 images from minival set.

The CityPersons dataset \cite{zhang2017citypersons} is a popular pedestrian detection benchmark, which is a challenging dataset for its large diversity in terms of countries and cities in Europe, seasons, person poses, and occlusion levels. The dataset has 2795, 1575, and 500 images for Train, Test, and Val subsets, respectively. It contains six object categories, namely, ignored region, person, group of people, sitter, rider, and other. All objects are classified into bare, reasonable, partially occluded and heavily occluded categories based on visibility of body part. The Reasonable (\textbf{R}) occlusion level subset contains pedestrian examples with visibility greater than 65\%, and height no less than 50 pixels for evaluation. The Heavy occlusion (\textbf{HO}) subset has visibility range from 20 to 65\% .

\subsubsection{Metrics.} Following the widely used evaluation protocol, generic object detection is measured by the metric of mean average precision (mAP) over all object categories, together with average recall (AR) for MS-COCO.

Only pedestrian examples taller than 50 pixels and within \textit{Reasonable} occlusion range are used for evaluating our approach. We report detection performance using log-average miss rate (MR) metric which is computed by averaging miss rates at nine False Positives Per Image (FPPI) rates evenly spaced between $10^{-2}$ to $10^0$ in log space \cite{dollar2011pedestrian}. 

\subsection{Implementation Details}
D-Wave quantum annealer currently supports over 2000 physical qubits however the minor embedding process takes nontrivial time when more than 45 qubits are fed for quantum processing. In order to reduce detection latency, raw detections for each object category is initially suppressed using NMS with predefined threshold and filtered by top detection scores, to ensure number of bounding boxes are within qubit upper bound of quantum processing we preset. Fortunately, such setting doesn't hurt real application much as it is rare to encounter the image case where there are over 45 objects of the same category.

\subsubsection{Training.} We base on the two-stage Faster R-CNN \cite{ren2015faster} baseline detector to implement quantum-soft QUBO suppression. For generic object detection, the detector is trained with backbone ResNet101 CNN architecture \cite{resnet} starting from a publicly available pretrained model for generic object detection datasets. We train the detector with mini-batch of 1 image using stochastic gradient descent with 0.9 momentum and 0.0005 weight decay on a single RTX 2080 Ti GPU. The learning rate is initially set to 0.001, and decays at a factor of 0.1 for every 5 epochs with 20 epochs in total for training. We set 7 epochs for large-scale MS-COCO training.

\subsubsection{Inference.} Our QSQS method only applies at suppression stage of testing phase. The greedy-NMS is replaced with Algorithm \ref{qsqs} where the weight parameters $w_1+w_2+w_3=1$ defined in Eq. \ref{e2} are tuned using pattern search algorithm \cite{psa}. After a group of searches, we fix them with $w_1 = 0.4$, $w_2=0.3$, and $w_3=0.3$ for the entire experiment. The original QUBO model is enhanced by introducing a score $s_i$ penalty term and an overlap $or(b_i, b_j)$ reward term if objectiveness score is smaller than confidence threshold. Sensitivity analysis is conducted for determining optimal pre-suppression NMS threshold before quantum processing. Bit string solution is retrieved from D-Wave annealer, then detection score above 0.01 after Gaussian rescoring function \cite{soft-nms} is applied to output final true locations by the detector.

\subsection{Ablation Study}
In this section, we conduct an ablative analysis to evaluate the performance of our proposed method on generic object detection datasets.

\subsubsection{Why quantum QUBO is enhanced?} Quantum-soft QUBO suppression is enhanced with three contributing factors, namely quantum QUBO (QQS), soft-NMS (QSQS), and adjustment terms (QSQS+enh), each of which either reduces inference latency or increases detection accuracy. Classical solver execution time is exponentially proportional to the number of binary variables to solve the NP-hard QUBO problem, while quantum annealing solves the larger problem in a divide-and-conquer paradigm by breaking it down to multiple sub-problems and finally produces a (near-)optimal solution \cite{shaydulin2019hybrid}.

Two potential drawbacks of QUBO suppression were pointed out in \cite{dpp}, i.e. blindness of equally selecting all detections and high penalty for occluded objects. However, these two issues can be efficiently solved by the enhanced model by introducing two adjustment factors for locations with detection cores lower than predefined objectiveness threshold. The penalty factor of 0.1 penalizes those lower-scored bounding boxes as they are likely to be false positives, while the reward factor of 0.7 reduces overlap between such lower-scored box and neighboring boxes accordingly. Impact of the enhanced model (QSQS+enh) with adjustment factors is shown in Table \ref{coco} on MS-COCO.

\begin{table*}[!ht]
\fontsize{10pt}{10pt}
\centering
\caption{Average precision (AP) results for MS-COCO test-dev set for various recall rates. The baseline Faster R-CNN uses standard NMS, while following three lines denote basic QUBO, soft-QUBO and enhanced QUBO, respectively. Best results are shown in bold. QSQS-enh shows no significant improvement over QSQS in MS-COCO.}

\begin{tabular}{|c| c| c| c| c| c| c| c| c| c| c|}
 \hline
\multicolumn{1}{|c|}{Method} & \multicolumn{1}{p{1.1cm}|}{AP 0.5:0.95} & \multicolumn{1}{p{1.1cm}|}{AP @0.5} & \multicolumn{1}{p{1.1cm}|}{AP small} & \multicolumn{1}{p{1.1cm}|}{AP medium} & \multicolumn{1}{p{1.1cm}|}{AP large} & \multicolumn{1}{p{1.1cm}|}{AR @10} & \multicolumn{1}{p{1.1cm}|}{AR @100}\\
 
 \hline  \hline 
 F-RCNN$^*$ \cite{ren2015faster} &  25.6 & 43.9 & 9.6 & 29.1 & 40.0 & 37.6 & 38.4 \\
 F-RCNN + QQS   &  21.8 & 35.4 & 6.6 & 24.3 & 35.0 & 27.0 & 27.0 \\
 F-RCNN + QSQS  &  25.7 & 44.2 & \textbf{9.8} & 29.2 & 40.3 & \textbf{38.5} & \textbf{39.4} \\
     F-RCNN + QSQS-enh  &  \textbf{25.8} & \textbf{44.3} & 9.7 & \textbf{29.3} & \textbf{40.4} & \textbf{38.5} & \textbf{39.4} \\
 
 \hline
 
  \multicolumn{8}{l}{F-RCNN$^*$ refers to F-RCNN \cite{ren2015faster} trained in the present study.}
  
 \end{tabular}
\label{coco}
\end{table*}

Soft-NMS has been adopted in few study and competition projects as it consistently generates better performance than greedy-NMS. The soft-QUBO idea in this paper is not the same as \cite{soft-nms}. However, the same rescoring function is applied to discard detections from D-Wave annealer. Impact of the soft model (QSQS) is also shown in Table \ref{coco} compared to basic QUBO model (QQS).

\subsubsection{Did quantum era arrive for object detection?} We may provide clues to this question from three perspectives, i.e. cost, accuracy, and inference latency of our QSQS method. The D-Wave Systems provides developers real-time quantum cloud service at commercial price of \$2000 per hour access to quantum processing unit. However, only problems with less than 45 qubits are mapped to quantum annealer for optimization in this study, and these small problems can be solved instantly such that the annealing process is undetectable and without time count. Since there is no extra QPU access cost, our method shares the same level cost with other existing methods.

Accuracy result shown in Table \ref{coco} indicates that QSQS surpasses the standard greedy-NMS embedded in most state-of-the-art detectors. More accuracy results are shown in Section 4.4.

The frame rate of QSQS is satisfying provided that Algorithm \ref{qsqs} is sort of more complicated than standard NMS. Even powered by quantum supremacy, extra inference latency caused by D-Wave cloud service connection, problem mapping, and annealing process is non-negligible. In order to better depict frame rate, we conduct the sensitivity analysis on inference latency vs. accuracy with varying initial overlap threshold $N_t$ (not lower than 0.3) and number of qubits (not higher than 45) on PASCAL VOC 2007. We choose initial overlap threshold from 0.3 to 0.7 with step size 0.1 and qubit upper bound from 15 to 45 with step size 5. Greedy-NMS achieves mAP of 74.20 (percent) over 20 object categories, and average inference latency is 65.87ms. QSQS achieves best performance with overlap threshold $N_t$ of 0.5 and 35 qubits upper bound.

\begin{figure*}
\centering
\includegraphics[width=12cm]{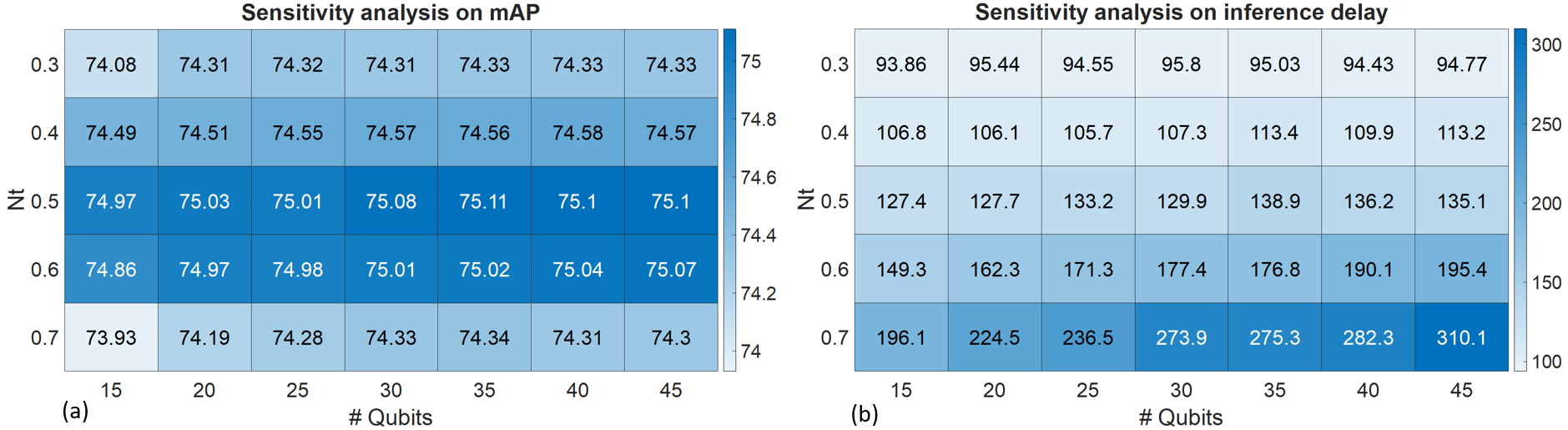}
\caption{Sensitivity analysis on detection accuracy mAP and average inference latency. QSQS achieves best accuracy with configuration of 0.5 initial overlap ratio and 35 qubit upper bound, at which the corresponding average inference latency is 138.9ms.}
\label{sensitivity}
\end{figure*}

The study results are shown in Fig. \ref{sensitivity}. Inference delay significantly relates to the initial overlap threshold, and the limit on number of qubits at higher $N_t$. Interestingly, QSQS underperforms at both high and low overlap threshold ends. Poor performance at low end 0.3 indicates that QSQS works better than greedy-NMS by introducing individual and correlation features. Performance at high end 0.7 indicates QSQS cannot handle well with large number of false positives. QSQS has performance gains with 28ms extra latency, and best performance with 73.02ms extra latency. It is also worth noting that we have submitted 223 problem instances with more than 45 qubits during this work, they are solved in 12ms on average by the D-Wave annealer. This indicates that the main latency overhead originates from initial false positive suppression and quantum cloud service connection.

The hybrid quantum-classical false positive suppression algorithm incurs no extra quantum computing cost, has better detection accuracy, and does not hurt much real-time detection application. In this sense, quantum era for object detection has already arrived.

\subsection{Results}
Parameters in our QSQS method are well tuned after sensitivity analysis. Its performance on generic object detection and pedestrian detection are compared with the state-of-the-art detectors on PASCAL VOC 2007 and CityPersons.

\subsubsection{Generic object detection.} 
\begin{figure*}
\centering
\includegraphics[width=12cm]{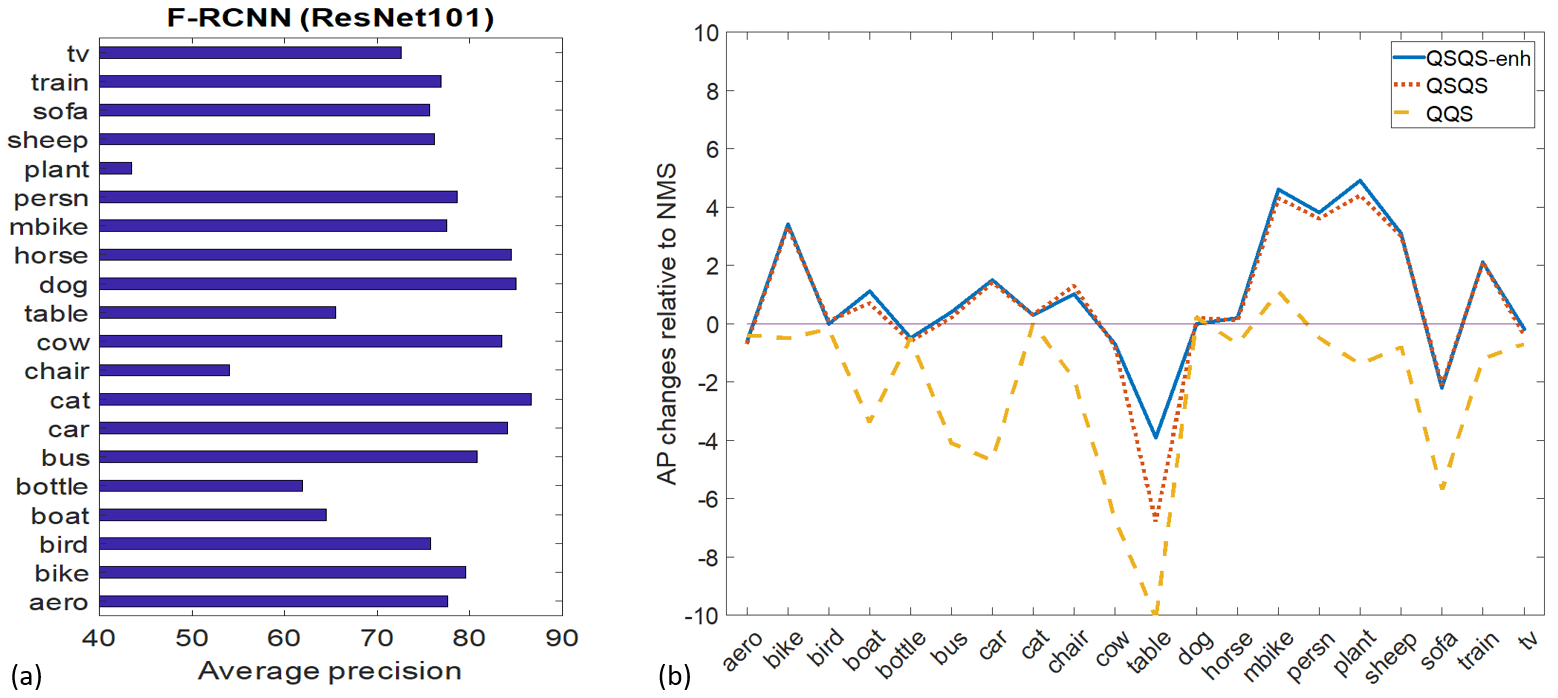}
\caption{(a) Baseline detector results for 20 object categories; (b) average precision changes for our three layer methods relative to the baseline. Curves above 0 show precision gains while that below 0 line show precision losses.}
\label{gen_od}
\end{figure*}

Faster R-CNN with QSQS is sufficiently trained on PASCAL VOC 2007 for fair comparison with greedy-NMS. The detection comparison over 20 object categories is summarized in Fig. \ref{gen_od}. It can be seen that performance increases a lot by adapting soft-NMS mechanism to basic QUBO model, and QSQS-enh performs slightly better only for few object categories. Interestingly, all our three schemes predict \texttt{table} and \texttt{sofa} significantly worse than standard NMS, but it becomes more accurate from QQS to QSQS-enh. Specifically, average precision for categories such as, \texttt{bike}, \texttt{motorbike}, and \texttt{person} are fairly improved by our method.

\begin{figure*}
\centering
\includegraphics[width=12cm]{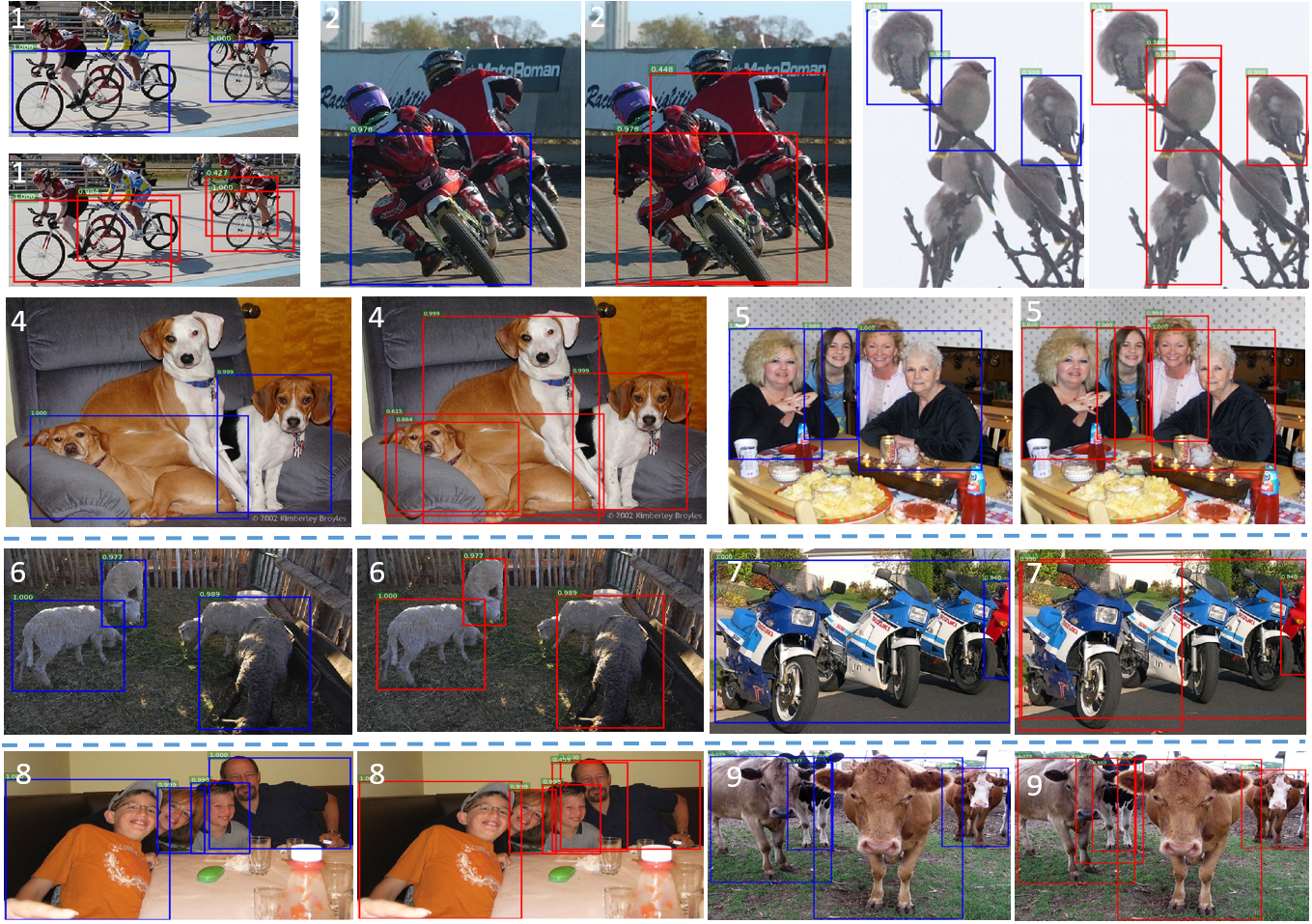}
\caption{Qualitative results. Image pairs in the top row shows cases where QSQS (in red boxes) performs better than NMS (in blue boxes). Images in bottom row show cases where NMS detects better. Pairs in the middle row are cases that cannot be properly detected by both methods due to poor raw detections.}
\label{qualitative}
\end{figure*}

Fig. \ref{qualitative} shows the qualitative results of generic object detections. Overlap threshold of 0.3 is set for greedy-NMS. Images 1-5 show examples where QSQS performs better than NMS, and images 8-9 show cases where NMS does better. Both do not detect well for images like 6-7. QSQS helps for images where false positives are partially overlapped with true detections, and it detects relatively better for image 3. False positive detection scores are fairly reduced for QSQS for images like 8 and 9. Bad detections in images 6 and 7 are essentially caused by proposed regions of interests, instead of post-processing schemes.

\begin{table*}
\fontsize{10pt}{10pt}
\centering
\caption{Best log-average Miss Rate for three schemes across 35 training epochs. Total evaluation time for 500 test images are shown for each scheme. Best values are shown in bold for each metric column.}
\begin{tabular}{|c| c| c| c| c| c| c|}
 \hline
   Method & Backbone & Time (s) & Reasonable & Bare & Partial & Heavy \\
 
 \hline  \hline 
 CSP$^*$ \cite{csp} + NMS & ResNet-50 & \textbf{83.63} & 27.48 & 20.58 & \textbf{27.34} & \textbf{65.78} \\
 CSP + Soft-NMS & ResNet-50 & 92.07 & 28.13 & 21.44 & 28.57 & 67.12 \\
 CSP + QSQS & ResNet-50 & 90.91 & \textbf{27.33} & \textbf{20.52} & 27.48 & 66.38 \\
 
 \hline
 \multicolumn{5}{l}{CSP$^*$ refers to CSP \cite{csp} trained in our own schedule.}
 \end{tabular}
\label{citypersons}
\end{table*}

\subsubsection{Pedestrian detection.} 
\begin{figure*}
\centering
\includegraphics[width=12cm]{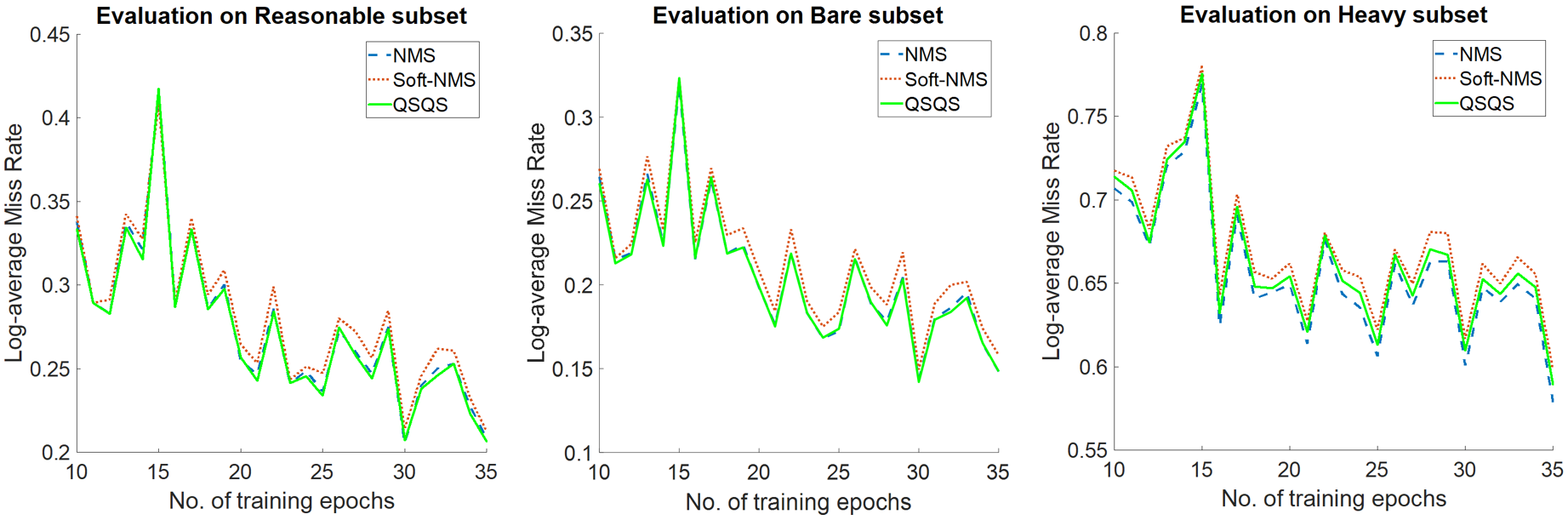}
\caption{Log-average Miss Rate comparison for three occlusion subsets across 35 training epochs. All three schemes show decreasing $MR^{-2}$ trends along increasing training epochs. QSQS outperforms NMS and soft-NMS for \textit{Reasonable} and \textit{Bare} subsets.}
\label{lamr}
\end{figure*}

We embed QSQS on Center and Scale Prediction (CSP) detector \cite{csp} as the enhanced post-processing scheme. CSP achieves best detection result in terms of log-average Miss Rate metric for two challenging pedestrian benchmark datasets. Table \ref{citypersons} and Fig. \ref{lamr} show $MR^{-2}$ comparison among NMS, Soft-NMS and QSQS on the more challenging CityPersons across 35 training epochs. Results are evaluated on standard \textit{Reasonable} subset, together with three other subsets with various occlusion levels. The best $MR^{-2}$ is displayed for each suppression scheme. It's worth noting that soft-NMS has the longest average inference delay (92.07s for 500 test images), and worst miss rates for all occlusion level subsets. In Fig. \ref{lamr}, QSQS shows least miss rates consistently among three schemes for \textit{Reasonable} and \textit{Bare} subsets across 35 training epochs. Standard NMS performs best for heavily occluded subset, which may indicate NMS detects better the objects with low visibility caused by other categories.

It's worth mentioning that QSQS is a false positive suppression scheme that performs well on both generic object detection and pedestrian detection application scenarios. The flexibility on composing score term of individual bounding box and correlation term between neighboring boxes entails its superiority of handling distinct tasks.

\section{Conclusion}
In this paper, we propose a novel hybrid quantum-classical QSQS algorithm for removing redundant object detections. The inspiration originates from basic QUBO suppression method, however it is impractical for real-world detection due to low accuracy and long inference latency. Detection accuracy in our work is improved by adapting soft-NMS method and introducing two adjustment factors for basic linear and quadratic terms. Furthermore, we leverage quantum advantage by solving QUBO problems (which are NP-hard) in D-Wave annealing system instead of classical solver that cannot solve such problems for real-time applications. The proposed algorithm incurs no extra QPU access cost.

We report the experimental results on generic object detection datasets, namely, MS-COCO and PASCAL VOC 2007, and pedestrian detection CityPersons datasets. The results show the proposed QSQS method improves  mAP from 74.20\% to 75.11\% for PASCAL VOC 2007 through two level enhancements. For benchmark pedestrian detection CityPersons, it consistently outperforms NMS and soft-NMS for bothe \textit{Reasonable} and \textit{Bare} subsets.

\section{Acknowledgment}
This work was supported in part by SRC (2847.001), NSF (CNS-1722557, CCF-1718474, DGE-1723687 and DGE-1821766), Institute for Computational and Data Sciences and Huck Institutes of the Life Sciences at Penn State.

\clearpage
%
%
\bibliographystyle{splncs04}
\bibliography{egbib}
\end{document}